\documentclass[conference]{IEEEtran}
\IEEEoverridecommandlockouts
\usepackage{cite}
\usepackage{amsmath}
\usepackage{amssymb}
\usepackage{bm}
\usepackage{graphicx}

\def\BibTeX{{\rm B\kern-.05em{\sc i\kern-.025em b}\kern-.08em
    T\kern-.1667em\lower.7ex\hbox{E}\kern-.125emX}}

\DeclareMathOperator{\vectorize}{vec}
\DeclareMathOperator{\tensorize}{ten}
\DeclareMathAlphabet\mathcalbf{OMS}{cmsy}{b}{n}
\newtheorem{remark}{Remark}

\begin{document}

\title{Recurrent Graph Tensor Networks: A Low-Complexity Framework for Modelling High-Dimensional Multi-Way Sequences}

\author{\IEEEauthorblockN{Yao Lei Xu, Danilo P. Mandic}
\IEEEauthorblockA{\textit{Department of Electrical and Electronic Engineering} \\
\textit{Imperial College London}\\
London, United Kingdom \\
\{yao.xu15,  d.mandic\}@imperial.ac.uk}
}

\maketitle

\begin{abstract}
Recurrent Neural Networks (RNNs) are among the most successful machine learning models for sequence modelling, but tend to suffer from an exponential increase in the number of parameters when dealing with large multidimensional data. To this end, we develop a multi-linear graph filter framework for approximating the modelling of hidden states in RNNs, which is embedded in a tensor network architecture to improve modelling power and reduce parameter complexity, resulting in a novel Recurrent Graph Tensor Network (RGTN). The proposed framework is validated through several multi-way sequence modelling tasks and benchmarked against traditional RNNs. By virtue of the domain aware information processing of graph filters and the expressive power of tensor networks, we show that the proposed RGTN is capable of not only out-performing standard RNNs, but also mitigating the Curse of Dimensionality associated with traditional RNNs, demonstrating superior properties in terms of performance and complexity. 
\end{abstract}


\section{Introduction}
\label{sec:intro}


Graphs and tensors have found numerous applications in deep learning systems. In this context, graph based methods have been used to generalize classical convolutional neural networks to irregular data domains, with graph neural networks achieving state-of-the-art results in a number of applications \cite{wu2020comprehensive}. On the other hand, tensor methods have been used to relax the computational complexity of neural networks \cite{Novikov2015tnn}, as well as to alleviate their notorious ``black-box" nature \cite{cohen2016expressive, calvi2019compression}. These promising results have also highlighted a void in literature regarding the combination of both techniques in order to solve deep learning challenges, especially in the area of sequence modelling. To this end, we introduce a novel Recurrent Graph Tensor Network (RGTN) framework for multi-way time-series modelling, which enhances the sequence modelling ability of Recurrent Neural Networks (RNNs) \cite{mandic2001recurrent} through tensor- and graph-theoretic concepts. 


The field of Graph Data Analytics (GDA) generalizes traditional signal processing concepts to irregular domains \cite{stankovic2020graphI, stankovic2020graphII, stankovic2020graphIII}, which are naturally represented as graphs. Developments in GDA have led to a range of spatial and spectral based techniques that generalize the notion of frequency and locality to irregular data, allowing for the processing of signals while taking into account the underlying data domain \cite{bronstein2017geometric}. Several concepts developed in GDA have found applications in deep learning, where graph filters can be implemented across multiple graph neural network layers to incorporate graph topology information \cite{wu2020comprehensive}.


Tensors are multi-linear generalization of vectors and matrices to multi-way arrays, which allows for a richer representation by not limiting the data to the classical ``flat-view" matrix approaches \cite{Cichocki2014}. Recent developments in tensor manipulation have led to Tensor Decomposition (TD) techniques that can represent high dimensional tensors through a contracting network of smaller core tensors. Such TD techniques can be used to compress the number of parameters needed to represent high-dimensional data, and have already found applications in deep learning. Notably, it has been shown that TD techniques, such as the Tensor-Train Decomposition (TTD) \cite{oseledets2011tensor}, can be used to compress neural networks considerably while maintaining comparable performance \cite{Novikov2015tnn, yang2017tensor, yu2017}.


However, despite promising results achieved in both individual fields, the full potential arising from the combination of graphs, tensors, and neural networks is yet to be explored, especially in the area of sequence modelling. To this end, we set out to investigate the extent to which a careful domain consideration of tensors and graphs can improve the complexity and performance of RNNs, by leveraging the theoretical frameworks underpinning graph machine learning and tensor networks. More specifically, we establish a novel structure for the modelling of RNN hidden states through a multi-linear graph filter embedded in a tensor network architecture, leading to a novel \textit{Recurrent Graph Tensor Network} (RGTN) framework. The so derived RGTN exploits both the ability of graphs to process data defined on irregular time-domains and the expressive power of tensor decomposition, resulting in a new class of expressive models with drastically lower complexity compared to standard RNNs. Our experimental results confirm the superiority of the proposed RGTN models, demonstrating desirable properties in terms of both performance and complexity across several multi-way sequence modelling tasks.


\newpage
\section{Theoretical Background}
\label{sec:theoretical_background}

\subsection{Spatial Graph Filters}
\label{sec:gsp}


A graph $\mathcal{G} = \{\mathcal{V}, \mathcal{E}\}$ is defined by a set of $N$ vertices (or nodes) $\textit{v}_n \subset \mathcal{V}$ for $n = 1, \ldots , N$, and a set of edges connecting the $n^{th}$ and $m^{th}$ vertices $\textit{e}_{nm} = (\textit{v}_n, \textit{v}_m) \in \mathcal{E}$, for $n=1,\ldots,N$ and $m=1,\ldots,N$. A signal on a given graph is defined by a vector $\textbf{f} \in \mathbb{R} ^ {N}$ such that $\textbf{f}: \mathcal{V} \rightarrow \mathbb{R}$, which associates a signal value to every node on the graph \cite{stankovic2020graphI}. 

A given graph can be fully described in terms of its weighted adjacency matrix, $\textbf{A} \in \mathbb{R} ^ {N \times N}$, such that $\textit{a}_{nm} > 0$ if $\textit{e}_{nm} \in \mathcal{E}$, and $\textit{a}_{nm} = 0$ if $\textit{e}_{nm} \notin \mathcal{E}$. The normalized weighted adjacency matrix is defined as $\tilde{\textbf{A}} = \textbf{D}^{-\frac{1}{2}} \textbf{A} \textbf{D}^{-\frac{1}{2}}$, where $\textbf{D} \in \mathbb{R} ^ {N \times N}$ is the diagonal degree matrix such that $d_{nn} = \sum_m \textit{a}_{nm}$ \cite{stankovic2020graphI}. The weighted adjacency matrix can be used as a shift operator to filter a set of $M$ signals on a graph with $N$ vertices, $\textbf{X} \in \mathbb{R}^{N \times M}$, as $\textbf{Y} = \sum_{k=0}^{K-1} \alpha_k \textbf{A}^k \textbf{X}$. Such a spatial graph filter represents a linear combination of vertex-shifted graph signals, which captures graph information at a local level \cite{stankovic2020graphII}. 

\subsection{Tensors and Tensor Networks}
\label{sec:tensors_and_TNs}

		
	
	
	
	
	

An order-$N$ tensor, $\mathcalbf{X}$ $\in \mathbb{R}^{I_1 \times I_2 \times \cdots \times I_N}$, represents an $N$-way array with $N$ modes, where the $n^{th}$ mode is of size $I_n$, for $n=1, 2, \dots, N$. Special instances of tensors include matrices ($\mathbf{X}  \in \mathbb{R}^{I_1 \times I_2}$), vectors ($\mathbf{x} \in \mathbb{R}^{I_1}$), and scalars ($x \in \mathbb{R}$), which are respectively tensors of order-2, 1, and 0. The $(i_1,i_2,\ldots,i_N)$ entry of a tensor is denoted by $x_{i_1i_2 \cdots i_N} \in \mathbb{R}$. A matrix can be \textit{reshaped} into a tensor through a process known as \textit{tensorization} \cite{Cichocki2014}, denoted by the operator $\tensorize(\cdot)$. A tensor can also be reshaped into a vector through the \textit{vectorization} process, denoted by the operator $\vectorize(\cdot)$. The tensor indices in this paper are grouped according to the Little-Endian convention \cite{Dolgov2014}.

An $(m,n)$-contraction, denoted by $\times^m_n$, between an $N$-th order tensor, $\mathcalbf{A} \in \mathbb{R}^{I_1\times \cdots \times I_n \times \cdots \times I_N}$, and an $M$-th order tensor, $\mathcalbf{B}\in \mathbb{R}^{J_1\times \dots \times J_m \times \dots \times J_M} $, with equal dimensions $I_n = J_m$, yields a tensor of order $(N+M-2)$, $\mathcalbf{C}\in \mathbb{R}^{I_1 \times \cdots \times I_{n-1} \times I_{n+1}  \times \cdots \times I_N \times J_1 \times \cdots \times J_{m-1} \times J_{m+1}  \times \cdots \times J_M}$, with entries defined as in (\ref{eq:cont}) \cite{Cichocki2014}. For the special case of matrices, $\textbf{A} \in \mathbb{R}^{I_1 \times I_2}$ and $\textbf{B} \in \mathbb{R}^{J_1 \times J_2}$ where $I_2=J_1$, the contraction, $\textbf{A} \times_2^1 \textbf{B}$, denotes the matrix multiplication, $\textbf{A}\textbf{B}$.
\begin{equation}\label{eq:cont}
	\begin{aligned}
		&c_{i_1 \dots i_{n-1} i_{n+1} \dots i_N j_1 \dots j_{m-1} j_{m+1} \dots j_M   } \\
		&= \sum_{i_n=1}^{I_n} a_{i_1 \dots i_{n-1} i_n i_{n+1} \dots i_N} b_{j_1 \dots j_{m-1} i_n j_{m+1} \dots j_M}   
	\end{aligned}
\end{equation}

A (left) Kronecker product between two tensors, $\mathcalbf{A} \in \mathbb{R}^{I_1 \times \cdots \times I_N}$ and $\mathcalbf{B} \in \mathbb{R}^{J_1 \times \cdots \times J_N}$, denoted by $\otimes$, yields a tensor of the same order, $\mathcalbf{C} \in \mathbb{R}^{I_1 J_1 \times \cdots \times I_N J_N}$, with entries $c_{\overline{i_1j_1},\ldots,\overline{i_Nj_N}} = a_{i_1 \ldots i_N} b_{j_1 \ldots j_N}$, where $\overline{i_n j_n} = j_n + (i_n - 1) J_n$ \cite{Cichocki2014}. For the special case of matrices $\textbf{A} \in \mathbb{R}^{I_1 \times I_2}$ and $\textbf{B} \in \mathbb{R}^{J_1 \times J_2}$, the Kronecker product yields a block-matrix:
\begin{equation}
    \textbf{A} \otimes \textbf{B} = 
    \begin{bmatrix}
    a_{i_1 i_2} \textbf{B} & \cdots & a_{i_1 I_2} \textbf{B} \\
    \vdots & \ddots & \vdots \\
    a_{I_1 i_2} \textbf{B} & \cdots & a_{I_1 I_2} \textbf{B} \\
    \end{bmatrix}
\end{equation}



A Tensor Network (TN) admits a graphical representation of tensor contractions, whereby each tensor is represented as a node, while the number of edges that extend from that node corresponds to the tensor order \cite{cichocki2016tensor}. An edge connecting two nodes represents a linear contraction over modes of equal dimensions between the connected tensors. 

\begin{figure}[t!]
	\centering
	\vspace{-3mm}
	\includegraphics[width=1\columnwidth]{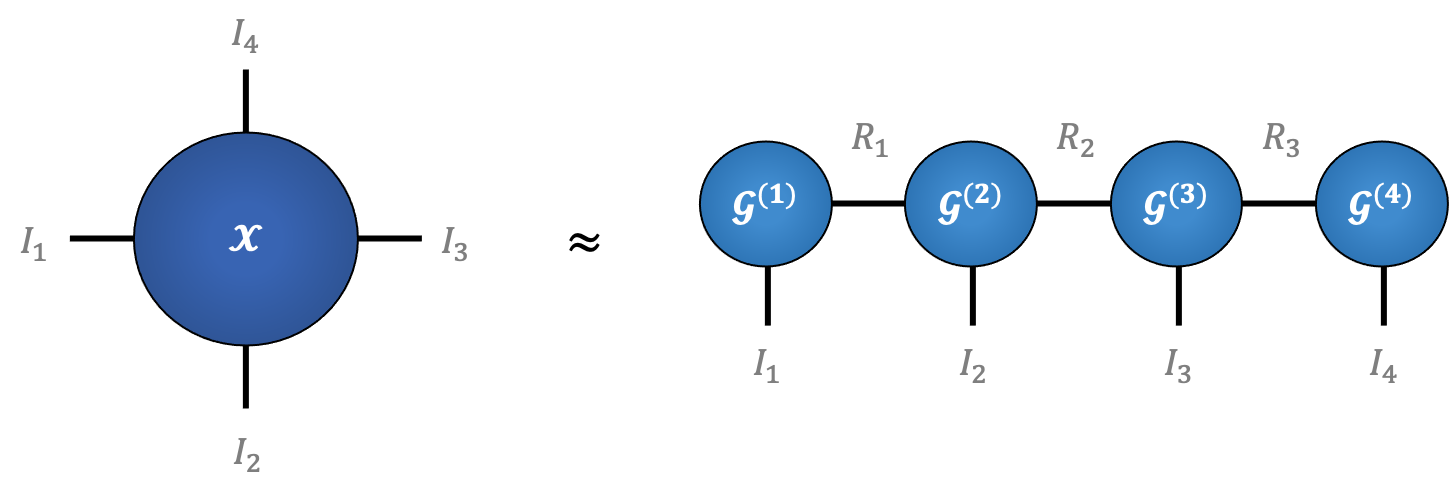}
	\vspace{-5mm}
	\caption{Tensor Network diagram of Tensor-Train Decomposition for an order-$4$ tensor, $\mathcalbf{X} \in \mathbb{R}^{I_1 \times I_2 \times I_3 \times I_4}$, according to (\ref{eq:TTContractions}). The dimensionality of the tensors are denoted in gray letters.}
	\label{fig:ttd_tn}
	\vspace{-3mm}
\end{figure}

Special instances of tensor networks include Tensor Decomposition (TD) networks. Such TD methods approximate high-order, large-dimensional tensors via contractions of smaller core tensors, which reduces the computational complexity drastically while preserving the data structure \cite{cichocki2016tensor, cichocki2017tensor}. For instance, the Tensor-Train (TT) decomposition \cite{Oseledets2009} \cite{oseledets2011tensor} is a highly efficient TD method that can decompose a large order-$N$ tensor, $\mathcalbf{X} \in \mathbb{R}^{I_1 \times I_2 \times \cdots \times I_N}$, into $N$ smaller core tensors, $\mathcalbf{G}^{(n)} \in \mathbb{R}^{ R_{n-1} \times  I_n \times R_n }$, as:
\begin{equation}\label{eq:TTContractions}
	\mathcalbf{X} = \mathcalbf{G}^{(1)} \times^1_2 \mathcalbf{G}^{(2)} \times^1_3 \mathcalbf{G}^{(3)} \times^1_3 \cdots \times^1_3 \mathcalbf{G}^{(N)}	
\end{equation}
where the set of $R_n$ for $n=0,\ldots,N$ and $R_0 = R_N = 1$ is referred to as the \textit{TT-rank}. By virtue of TT, the number of entries in the original tensor is drastically reduced from an exponential $\prod_{n=1}^N I_n$ to a linear $\sum_{n=1}^N R_{n-1} I_n R_{n}$ in the dimensions $I_n$, which is highly efficient for high $N$ and low TT-rank. An illustration of TT decomposition in TN notation is provided in Figure \ref{fig:ttd_tn}.

\subsection{Recurrent Neural Networks}

Recurrent Neural Networks (RNNs) \cite{mandic2001recurrent} \cite{khalifa2020bio} are among the most successful deep learning tools for sequence modelling. A standard RNN layer captures time-varying dependencies by processing hidden states, $\mathbf{h}_t \in \mathbb{R}^M$, at time $t$ through \textit{feedback} (or \textit{recurrent}) weights as:
\begin{equation}    
    \label{eq:RNN_unit_ht}
    \mathbf{h}_t = \sigma_h (\mathbf{W}^{(h)} \mathbf{h}_{t-1} + \mathbf{W}^{(x)} \mathbf{x}_t + \textbf{b}^{(h)})
\end{equation} 
where $\mathbf{h}_{t-1} \in \mathbb{R}^M$ is the hidden state vector from the previous time-step, $\mathbf{x}_t \in \mathbb{R}^N$ is the input features vector at time $t$, $\mathbf{W}^{(h)} \in \mathbb{R}^{M \times M}$ is the feedback matrix, $\mathbf{W}^{(x)} \in \mathbb{R}^{M \times N}$ is the input weight matrix, $\mathbf{b}^{(h)} \in \mathbb{R}^M$ is an optional bias vector, and $\sigma_h(\cdot)$ is an optional element-wise activation function.

Finally, after extracting the hidden states, these can be passed through additional weight matrices to generate outputs, $\textbf{y}_t \in \mathbb{R}^{P}$ at time $t$, in the form:
\begin{equation}    
    \label{eq:RNN_unit_yt}
    \mathbf{y}_t = \sigma_y (\mathbf{W}^{(y)} \mathbf{h}_{t} + \textbf{b}^{(y)})
\end{equation} 
where $\mathbf{W}^{(y)} \in \mathbb{R}^{P \times M}$ is the output weight matrix, $\mathbf{h}_{t}$ is the hidden state at time $t$, $\textbf{b}^{(y)}$ is an optional bias vector, and $\sigma_y(\cdot)$ is an optional element-wise activation function.

\newpage
\section{RECURRENT GRAPH TENSOR NETWORKS}
\label{sec:tensor_network_model} 

\subsection{General Recurrent Graph Tensor Networks}
\label{sec:general_graph_conv_for_RNN}

Consider the RNN forward pass in (\ref{eq:RNN_unit_ht}) without the optional bias vector and activation function:
\begin{equation} \label{eq:simple_rnn_eq_lin}
    \mathbf{h}_t = \mathbf{W}^{(h)} \mathbf{h}_{t-1} + \mathbf{W}^{(x)} \mathbf{x}_t
\end{equation}
Denote $\hat{\textbf{x}}_t = \textbf{W}^{(x)}\textbf{x}_t \in \mathbb{R}^M$, for $t=1,\ldots,\tau$ time-steps; then (\ref{eq:simple_rnn_eq_lin}) can be written in a block-matrix form: 
\vspace{-1mm}
\begin{equation}
\label{eq:BlockMatrixRecurrency}
{\small
\begin{bmatrix}
\mathbf{h}_{\tau} \\
\mathbf{h}_{\tau-1} \\
\vdots \\
\mathbf{h}_1
\end{bmatrix}
 = 
\begin{bmatrix}
{(\mathbf{W}^{(h)})}^{0} & {(\mathbf{W}^{(h)})}^{1} & \cdots & {(\mathbf{W}^{(h)})}^{\tau-1}\\
0 & {(\mathbf{W}^{(h)})}^{0} & \cdots & {(\mathbf{W}^{(h)})}^{\tau-2}\\
\vdots & \vdots & \ddots & \vdots \\
0 & 0 & \cdots & {(\mathbf{W}^{(h)})}^{0}\\
\end{bmatrix}
\begin{bmatrix}
\hat{\textbf{x}}_{\tau} \\
\hat{\textbf{x}}_{\tau-1} \\
\vdots \\
\hat{\textbf{x}}_1
\end{bmatrix}
}
\end{equation}

We now define: (i) ${\textbf{X}} \in \mathbb{R} ^ {\tau \times N}$, as the input matrix generated by stacking row-vectors, ${\textbf{x}}_t$, over $\tau$ successive time-steps; (ii) $\hat{\textbf{X}} \in \mathbb{R} ^ {\tau \times M}$, as $\hat{\textbf{X}} = \textbf{X} \times_2^2 \textbf{W}^{(x)}$; (iii) $\textbf{H} \in \mathbb{R} ^ {\tau \times M}$, as the matrix generated by stacking hidden state vectors, $\textbf{h}_t$, as row-vectors over $\tau$ time-steps; and (iv) $\textbf{R} \in \mathbb{R}^{\tau M \times \tau M}$, as the block matrix composed by the powers of $\textbf{W}^{(h)}$ from (\ref{eq:BlockMatrixRecurrency}). This allows (\ref{eq:BlockMatrixRecurrency}) to be expressed compactly as: 
\begin{equation} \label{eq:hidden_state_pass_1}
\vectorize(\textbf{H}) = \textbf{R} \times_2^1 \vectorize(\hat{\textbf{X}})
\end{equation}

Without loss of generality, we shall further restrict the feedback matrix, $\textbf{W}^{(h)}$, to be a scaled idempotent matrix, that is $\textbf{W}^{(h)} = c \textbf{W}^{(r)}$, where $c$ is a positive scaling constant strictly less than 1, and $\textbf{W}^{(r)}$ is an idempotent matrix that models how information propagates between successive time-steps. For this setup, the feedback matrix has the property ${(\textbf{W}^{(h)})}^p = c^p {\textbf{W}^{(r)}}$, for $p>0$. This allows the block matrix $\textbf{R}$ to be decomposed as: 
\begin{equation}
    \label{eq:R_decomposition}
    \textbf{R} = \textbf{I} + \textbf{A} \otimes \textbf{W}^{(r)}
\end{equation}
where $\textbf{A} \in \mathbb{R}^{\tau \times \tau}$ contains the constants $c^p$, as:
\begin{equation}
{
\label{eq:time_adj_matrix}
\textbf{A}
 = 
\begin{bmatrix}
0 & {c}^{1} & \cdots & {c}^{\tau-1}\\
0 & 0 & \cdots & {c}^{\tau-2}\\
\vdots & \vdots & \ddots & \vdots \\
0 & 0 & \cdots & 0\\
\end{bmatrix}
}
\end{equation}
Note that the matrix, $\textbf{A}$, can be interpreted as the weighted graph adjacency matrix connecting $\tau$ successive time-steps as vertices (nodes). This also justifies its triangular (directed) nature, since only past information can influence future states but not vice-versa.

We now denote, $\mathcalbf{R} \in \mathbb{R}^{\tau \times M \times \tau \times M}$, as the $4$-th order tensorization of $\textbf{R}$, that is $\mathcalbf{R} = \tensorize(\textbf{I} + \textbf{A} \otimes \textbf{W}^{(r)})$; this allows us to simplify the expression in (\ref{eq:hidden_state_pass_1}) via a double tensor contraction, and express the general Recurrent Graph Tensor Network filtering operation in its complete form as:
\vspace{-1mm}
\begin{equation}
\label{eq:hidden_states_double_contraction}
\textbf{H} = \mathcalbf{R} \times_{3,4}^{1,2} \textbf{X} \times_2^2 \textbf{W}^{(x)}
\vspace{-1mm}
\end{equation}

\normalfont The proposed filtering operation in (\ref{eq:hidden_states_double_contraction}) can be used to extract features from time-series data, \textbf{X}, in a neural network. We will refer to such neural network models as general Recurrent Graph Tensor Networks (gRGTN).

\subsection{Simplified Recurrent Graph Tensor Networks}
\label{sec:special_graph_conv_for_RNN}

To establish a link between the proposed RGTN filtering operation and classical spatial graph filters, we shall now consider a special case of equation (\ref{eq:hidden_states_double_contraction}).

Consider a special case where $\textbf{W}^{(r)} = \textbf{I}$. This implies that $\mathbf{W}^{(h)} = c \textbf{I}$, which simplifies the hidden state evolution in (\ref{eq:simple_rnn_eq_lin}) as $\mathbf{h}_t = c \mathbf{h}_{t-1} + \hat{\textbf{x}}_t$. This corresponds a simplified system model where the past information is propagated to the future with a scaling constant of $c$. This simplifies (\ref{eq:hidden_states_double_contraction}) as:
\begin{equation}\label{eq:derive_simplified_rgf}
	\begin{aligned}
		\textbf{H} &= \mathcalbf{R} \times_{3,4}^{1,2} \textbf{X} \times_2^2 \textbf{W}^{(x)} \\
		&= \tensorize(\textbf{I} + \textbf{A} \otimes \textbf{W}^{(r)}) \times_{3,4}^{1,2} ( \textbf{X} \times_2^2 \textbf{W}^{(x)}) \\
		&= \tensorize(\textbf{I} + \textbf{A} \otimes \textbf{I}) \times_{3,4}^{1,2} (\textbf{X} \times_2^2 \textbf{W}^{(x)}) \\
		&= (\textbf{I} + \textbf{A}) \times_{2}^{1} (\textbf{X} \times_2^2 \textbf{W}^{(x)}) \\
		&= (\textbf{I} + \textbf{A}) \times_2^1 \hat{\textbf{X}}
	\end{aligned}
\end{equation}

Notice that (\ref{eq:derive_simplified_rgf}) is equivalent to $\textbf{H} = \sum_{k=0}^{K-1} \alpha_k \textbf{A}^k \hat{\textbf{X}}$, 
which is precisely a spatial graph filter as discussed in Section \ref{sec:gsp}, where $K=2$, $\alpha_k=1$, and $\textbf{A}$ is the weighted graph adjacency matrix that enforces the directed flow of time. We will refer to neural networks employing equation (\ref{eq:derive_simplified_rgf}) for feature extraction as simplified Recurrent Graph Tensor Networks (sRGTN). 

\begin{table*}[t!]
\centering
\caption{{Experiment Data Modalities}}
\label{tab:data_modes}
\begin{tabular}{l|ll|ll|ll|}
\cline{2-7}
                                                      & Mode 1        &           & Mode 2        &           & Mode 3               &           \\
                                                      & Physical Mode & Dimension & Physical Mode & Dimension & Physical Mode        & Dimension \\ \hline
\multicolumn{1}{|l|}{Air Quality Forecasting}   & Time          & 6         & Site          & 12        & Air Quality Features & 27        \\ \cline{2-7} 
\multicolumn{1}{|l|}{Temperature Forecasting}   & Time          & 6         & City          & 14        & Temperature Features & 4         \\ \cline{2-7} 
\multicolumn{1}{|l|}{House Price Forecasting}   & Time          & 6         & House Type    & 4         & Price Index Features & 2         \\ \cline{2-7} 
\multicolumn{1}{|l|}{Activity Recognition} & Time          & 24        & Sensor        & 3         & Measurement Features & 3         \\ \hline
\end{tabular}
\end{table*}

\subsection{Tensor Network Formulation}
\label{sec:tensor_network_formulation}

\begin{figure}[h!]
	\centering
	\includegraphics[width=1\columnwidth]{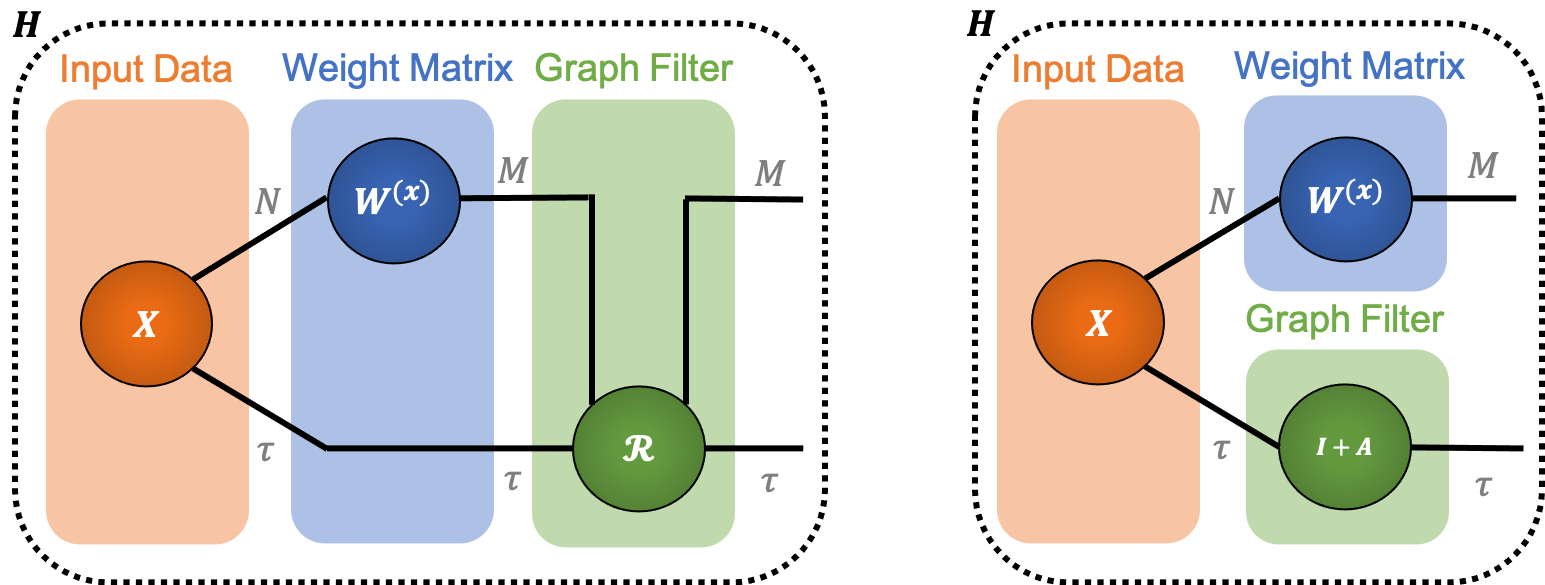}
	\caption{{Tensor Network (TN) diagram of the gRGTN filtering operation (left) according to (\ref{eq:hidden_states_double_contraction}), and the sRGTN filtering operation (right) according to (\ref{eq:derive_simplified_rgf}). The nodes of the TN diagram represent different tensors, while the edges represent tensor contractions over common dimensions between tensors. The dimensions of different tensors are denoted in gray letters.}}
	\label{fig:filteringTNs}
\end{figure}

Consider the gRGTN filtering operation in (\ref{eq:hidden_states_double_contraction}). The multi-linear nature of the tensor $\mathcalbf{R}$ and the associated double tensor contraction naturally admits a Tensor Network (TN) representation, as shown in Figure \ref{fig:filteringTNs} (left). Similarly, the sRGTN filtering in (\ref{eq:derive_simplified_rgf}) also admits a TN representation with a simpler topology, as shown in Figure \ref{fig:filteringTNs} (right). This allows the hidden state modelling operation to benefit from the enhanced expressive power of tensors, which are not limited to the standard ``flat-view" matrix methods \cite{cichocki2016tensor, cichocki2017tensor}.

\begin{figure}[h!]
	\centering
	\includegraphics[width=0.8\columnwidth]{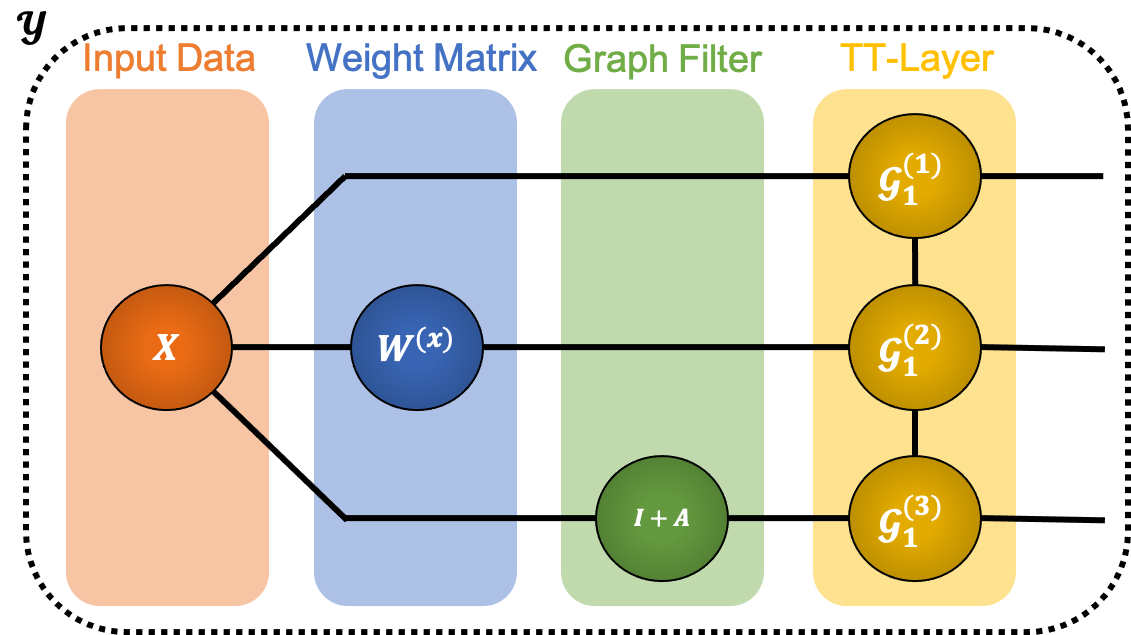}
	\caption{{A gRGTN model (top) and a sRGTN model (bottom) designed to handle order-3 input time-series tensors. In addition to the proposed filtering operations, it leverages on the power of Tensor-Train (TT) decomposition networks (in yellow) to achieve high expressive power at low parameter complexity, which is inherently compatible with the multi-way nature of the RGTN framework.}}
	\label{fig:exp_net_simp}
\end{figure}

By integrating the concept of graph filtering in a TN framework, we can easily design network architectures for processing time-series data of any modalities, as well as leverage on the power of tensor decomposition to boost its expressive power while maintaining low complexity. For illustration, Figure \ref{fig:exp_net_simp} shows TN models designed to process multi-way time series data as order-3 input tensors (i.e. the time-series features are indexed along a time-mode and an additional physical mode), which uses appropriate Tensor-Train (TT) networks to process filtered multi-way time-series data.

\begin{figure}[h!]
	\centering
	\includegraphics[width=0.8\columnwidth]{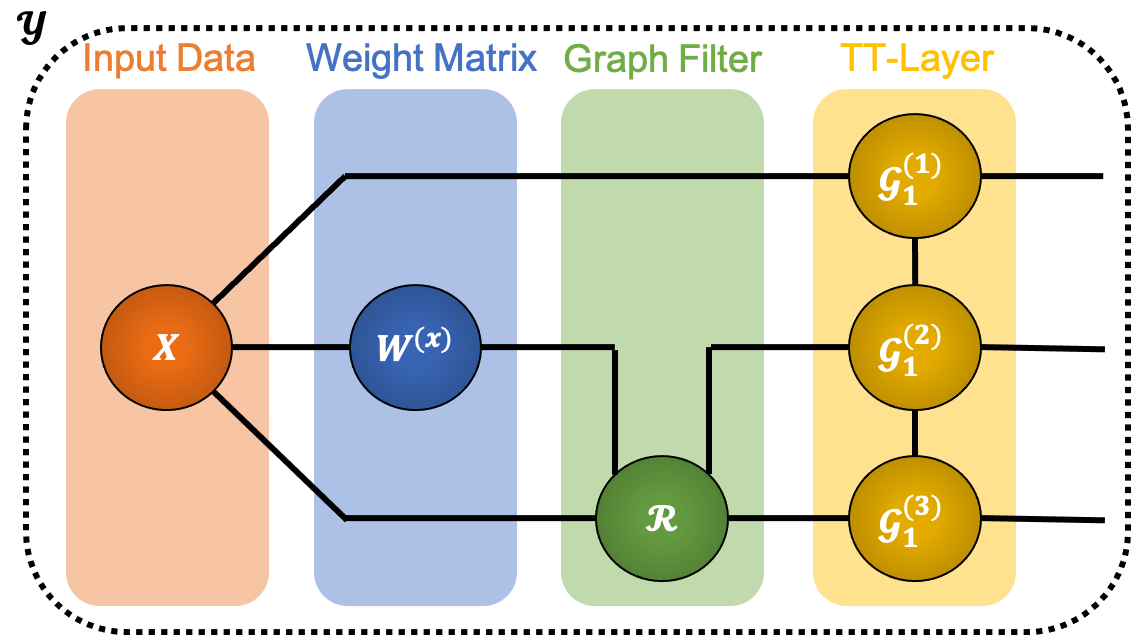}
	\label{fig:exp_net}
	\vspace{-2mm}
\end{figure}

\begin{remark}
    \normalfont The double tensor contraction with $\mathcalbf{R}$ in gRGTN implies a stronger coupling of features with the underlying time-domain represented in graph form, thus yielding enhanced expressive power over the decoupled contractions in sRGTN.
\end{remark}

\begin{remark}
    \normalfont The Tensor-Train layers in Figure \ref{fig:exp_net_simp} can be interpreted as tensorized fully-connected neural network layers compressed via Tensor-Train decomposition, which drastically reduces the number of parameters required to achieve the same expressive power \cite{Novikov2015tnn, yang2017tensor}.
\end{remark}

\section{EXPERIMENTS}
\label{sec:results}

\subsection{Datasets}

To validate the expressive power of the proposed gRGTN and sRGTN models, we verified their performance in a number of multi-way time-series modelling tasks, including: 
\begin{enumerate}
    \item  \textit{Beijing Multi-Site Air Quality Forecasting} \cite{zhang2017cautionary}. This dataset consists of various air quality measurements obtained across 12 different sites in China recorded at an hourly rate. The learning task for this dataset is to forecast the air quality level across all 12 sites in the next hour. 
    \item \textit{Global Land Temperature Forecasting} \cite{rohde2013new}. This dataset consists of monthly temperature recordings obtained across multiple cities around the world. The learning task for this dataset is to forecast the average temperature across 14 major cities in India during the next month. 
    \item \textit{Liverpool House Price Forecasting} \cite{uk_hpi}. This dataset consists of monthly price indices across 4 different types of houses in Liverpool, United Kingdom. The learning task for this dataset is to forecast the price indices of different house types in the next month.
    \item \textit{Multi-Sensor Activity Recognition} \cite{palumbo2016human}. This dataset consists of multi-sensor measurements of human bodies when performing different physical activities. The learning task for this dataset is to classify the physical activity from the multi-sensor measurements.
\end{enumerate}
All of the considered data are multi-modal time-series tensors of order-3. The exact modalities of the input data tensors are summarized in Table \ref{tab:data_modes}.

\subsection{Benchmark Models and Metrics}

We compared the performance of the proposed gRGTN and sRGTN models against standard RNN, GRU, and LSTM based neural networks. For comparable results, all models have the exact same model architecture, hidden units, activation functions, and training method, with the only differences being: (i) the feature extraction layer used, which can be based on gRGTN, sRGTN, RNN, GRU, or LSTM, and (ii) the fully-connected dense layers, which are replaced by the equivalent TT networks for gRGTN and sRGTN as shown in Figure \ref{fig:exp_net_simp} \cite{Novikov2015tnn}. For more details, please refer to the full experiment code provided on GitHub\footnote{The code is available on www.github.com/gylx/RGTN}.

We compared the considered models across the proposed experiments both in terms of performance and complexity. In terms of performance metrics, we used out-of-sample Mean Absolute Error (MAE) for the regressions tasks related to datasets (1), (2), and (3), and classification accuracy for the classification task related to dataset (4). In terms of complexity, we compare the number of trainable parameters needed to achieve the same model specifications.

\begin{table*}[t]
\centering
\caption{{Performance and Complexity of the Considered Models.}}
\begin{tabular}{|l|lllll|}
\hline
  Test Set Score               & gRGTN             & sRGTN    & RNN              & GRU      & LSTM             \\ \hline
Air Quality Forecasting (MAE)   & \textbf{0.01598}  & 0.01742  & 0.01872          & 0.01706  & 0.01652          \\
Temperature Forecasting (MAE)   & 0.20959           & 0.27491  & 0.19905          & 0.18050  & \textbf{0.17744} \\
House Price Forecasting (MAE)   & 0.72946           & 0.76768  & \textbf{0.71195} & 0.74081  & 0.73463          \\
Activity Classification (Accuracy) & \textbf{79.883\%} & 78.740\% & 50.731\%         & 79.398\% & 78.629\%         \\ \hline
\end{tabular}
\label{tab:exp_res}
\end{table*}

\begin{table*}[t]
\centering
\begin{tabular}{|l|lllll|}
\hline
Number of Trainable Parameters  & gRGTN             & sRGTN    & RNN              & GRU      & LSTM             \\ \hline
Air Quality Forecasting   & 556  & \textbf{492}  & 2844  & 8196  & 10836          \\
Temperature Forecasting   & 406  & \textbf{342}  & 718   & 1782  & 2278 \\
House Price Forecasting   & 220  & \textbf{156}  & 244   & 540   & 652          \\
Activity Classification   & 301  & \textbf{237}  & 261   & 573   & 693         \\ \hline
\end{tabular}
\end{table*}

\subsection{Experiment Results}

The experiment results are summarized in Table \ref{tab:exp_res}. The top table shows the test set performance for three regression tasks (measured in MAE) and one classification task (measured in accuracy) achieved by the considered models. The bottom table shows the corresponding number of trainable parameters needed for each task. 

By virtue of its graph and tensor structure, the proposed gRGTN model achieved the best performance overall, obtaining the highest score for 2 out of 4 datasets, while using drastically less number of trainable parameters compared to standard RNN, GRU, and LSTM models. On the other hand, the sRGTN model achieved the lowest parameter complexity due to its approximation assumption of $\textbf{W}^{(r)} = \textbf{I}$, but at the cost of marginally reduced performance.


\section{CONCLUSION}
\label{sec:conclusion}
We have introduced a novel Recurrent Graph Tensor Network (RGTN) framework for modelling time-series data, by combining the expressive power of tensor networks with the ability of graphs to account for the structure underlying time-series data. Experiment results have verified the desirable properties of the proposed RGTN framework, which outperformed standard RNN, GRU, and LSTM models across multiple time-series modelling tasks, and at a drastically reduced parameter complexity.

\bibliographystyle{IEEEtran.bst}
\bibliography{references.bib}

\end{document}